\documentclass[10pt,twocolumn,letterpaper]{article}

\usepackage[algorithms]{wacv} 
\usepackage{graphicx}
\usepackage{booktabs}

\usepackage{multirow}
\newcommand{\rot}[1]{\rotatebox{45}{\makecell{#1}}}
\usepackage{makecell}
\usepackage{threeparttable}
\usepackage{xcolor}
\usepackage{pifont}
\newcommand{\cmark}{\textcolor{green!60!black}{\ding{51}}}
\newcommand{\xmark}{\textcolor{red}{\ding{55}}}
\definecolor{goodgreen}{RGB}{40,140,60}
\definecolor{badred}{RGB}{180,60,60}

\usepackage{tikz}

\usepackage[accsupp]{axessibility} 
\usepackage{comment}

\definecolor{wacvblue}{rgb}{0.21,0.49,0.74}
\usepackage[pagebackref,breaklinks,colorlinks,allcolors=wacvblue]{hyperref}

\def\wacvPaperID{134} 
\def\confName{WACV}
\def\confYear{2027}

\title{From Explicit References to Scene Manifolds: Distributional Fidelity and Realism for Radiance Field Quality Assessment}

\author{
Saeed Mahmoudpour$^{1,2}$ \quad
Gi-Mun Um$^{3}$ \quad
Hyon-Gon Choo$^{3}$ \quad
Peter Schelkens$^{1,2}$\\
$^{1}$Vrije Universiteit Brussel, Department of Electronics and Informatics, Belgium\\
$^{2}$imec, Kapeldreef 75, B-3001 Leuven, Belgium\\
$^{3}$Electronics and Telecommunications Research Institute (ETRI), Republic of Korea\\
{\tt\small
\{saeed.mahmoudpour,peter.schelkens\}@vub.be,
\{gmum,hyongonchoo\}@etri.re.kr
}
}
\begin{document}
\maketitle
\begin{abstract}
Radiance field representations such as 3D Gaussian Splatting (3DGS) enable high-quality novel view synthesis but can introduce complex, view-dependent artifacts from reconstruction, rendering, and compression.~Reliable perceptual quality assessment (QA) is thus essential for evaluating rendered views and guiding the design of perceptually faithful scene representations.~Existing full-reference QA metrics require an aligned reference image, while recent cross-reference metrics relax this requirement by comparing a test view with non-aligned references.~However, under wide-baseline radiance field settings, selecting a reliable nearby reference can be difficult, particularly when evaluating views along arbitrary trajectories and poses.~We propose SCODA, a lightweight scene-conditioned objective QA method that shifts QA from explicit image-to-image comparison to scene-manifold modeling.~High-quality observations of each scene are represented as a multivariate Gaussian distribution in deep feature space, producing a semantic fidelity score that measures deviation from the scene distribution.~A weakly-supervised distortion-aware patch discriminator provides a complementary realism signal, and both cues are combined through an unsupervised bounded fusion strategy.~Experiments on multiple benchmarks show strong agreement with human judgments and robust generalization across GS- and NeRF-generated views and trajectories. Code is publicly available at \url{https://gitlab.com/saeedmp/scoda}.

\end{abstract}
    
\section{Introduction}
\label{sec:intro}

Modeling 3D real-world scenes and rendering them from novel viewpoints has long been a central problem in computer vision \cite{hanrahan1996light, hedman2018deep}.~Over the past decades, this problem has evolved from explicit multi-view depth estimation and image-based rendering techniques \cite{chaurasia2013depth, goesele2007multi} toward recent learning-based representations that model complex scenes in a continuous function space, such as Neural Radiance Fields (NeRF) \cite{mildenhall2021nerf} and 3D Gaussian Splatting (3DGS) \cite{kerbl20233d}.~These representations enable high-quality novel-view synthesis and dense appearance modeling across viewpoints, capturing fine-grained visual details and rendering real-world environments with unprecedented visual quality. However, these advantages come at a significant cost in storage. In particular, 3DGS maintains millions of parameters describing spatial positions, covariance structures, opacity values, and appearance attributes, leading to a substantial memory footprint. Consequently, compression and rate–distortion optimization become essential for practical deployment, motivating recent efforts toward more compact and efficient representations \cite{bagdasarian24073dgs, morgenstern2024compact}.

\begin{figure*}[t]
\centering

\includegraphics[width=0.95\linewidth]{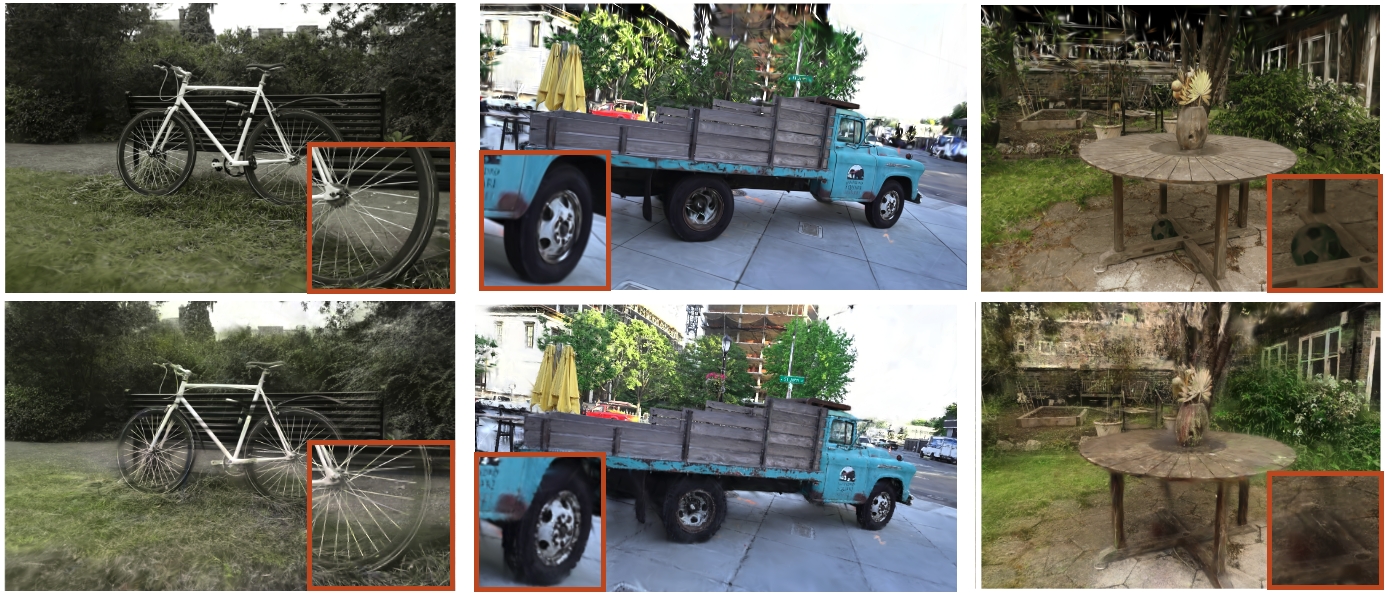}

\vspace{1mm}

\resizebox{0.80\linewidth}{!}{
\begin{tabular}{l|cccccccccc|cc}
\toprule
\textbf{Reference view}
& PSNR & SSIM & VIF & FSIM & LPIPS & DISTS & DeepDC
& PuzzleSim & CrossScore & NOVA
& \textbf{SCODA (Ours)} & \textbf{Human} \\
\midrule
Aligned
& \xmark & \xmark & \xmark & \xmark & \xmark
& \cmark & \cmark & \cmark & \cmark & \cmark
& \cmark & \cmark \\

Nearby non-aligned
& \xmark & \xmark & \xmark & \xmark & \xmark
& \xmark & \cmark & \xmark & \cmark & \cmark
& \cmark & \cmark \\

Distant non-aligned
& \xmark & \xmark & \xmark & \xmark & \xmark
& \xmark & \xmark & \xmark & \xmark & \xmark
& \cmark & \cmark \\
\bottomrule
\end{tabular}
}

\caption{Failure cases of quality metrics on GS artifacts. In each image pair, the first-row rendering is preferred by human observers over the second-row rendering. A check mark indicates agreement with human preference, while a cross mark indicates disagreement. We consider three reference-availability regimes: aligned reference, where the reference view matches the test pose; nearby non-aligned reference, where same-scene reference views are nearby but not pixel-aligned; and distant non-aligned reference, where same-scene views have limited overlap and stronger viewpoint changes. Our scene-manifold formulation remains consistent with human preference.}
\label{fig:teaser}
\end{figure*}

Despite their rendering quality, radiance field representations can produce artifacts that differ substantially from conventional 2D image degradations. These artifacts may arise from imperfect scene reconstruction, limitations of the underlying representation, rendering approximations, or subsequent compression of the scene model. In 3DGS, errors in Gaussian geometry, opacity, or appearance can cause misalignment, floaters, transparency artifacts, color shifts, and view-dependent flickering. Since these artifacts arise from a structured 3D representation, they may also propagate across viewpoints. These characteristics complicate visual quality assessment (QA) and reliable methods are essential not only for benchmarking rendered views, but also for guiding reconstruction, rendering, and compression methods toward better perceptual quality.

Objective QA methods are computational models to predict visual quality aligned with human judgment. They are commonly classified as full-reference (FR), no-reference (NR), or, more recently for novel view synthesis, cross-reference (CR) methods.~FR methods compare a test image with a pristine reference captured at the same camera pose.~In practice, synthesized views are commonly evaluated using conventional FR metrics such as PSNR and SSIM~\cite{ssim}. However, increasing evidence indicates that they are often inconsistent with human judgments~\cite{martin2025gs, xing20253dgs2}. The use of FR metrics is also constrained by the sparsity of reference camera poses, which limits evaluation to viewpoints where ground-truth images are available.

Recent CR methods~\cite{wang2024crossscore, hermann2025puzzle, ghildyal_nova_2026} relax this requirement by comparing a rendered view with non-aligned, nearby reference views. Nevertheless, they still require selecting one or a set of explicit reference views, with sufficient visual overlap or nearby viewpoint coverage. This becomes a bottleneck when evaluating radiance field models, where quality should be assessed along arbitrary trajectories and query poses across the scene. In such wide-baseline settings, captured views can be weakly overlapping, and the definition of a usable reference becomes ambiguous.~This motivates a scene-conditioned formulation that compares each rendering to a scene-level reference representation rather than to a selected reference view.~NR approaches~\cite{mittal2012making, chen2024topiq} bypass the need for explicit references, but typically achieve lower accuracy without retraining and often struggle to generalize across diverse scenes and datasets. 

Motivated by this reference availability bottleneck, we shift from explicit reference-view comparison to scene-conditioned distribution modeling. Rather than seeking a matched or neighboring reference image, our approach constructs a reference scene distribution from high-quality observations of the scene and evaluates each novel view by its deviation from this scene-level representation.~This formulation produces a strong semantic fidelity signal that remains reference-aware while avoiding the need for an explicit reference image.~Our fidelity branch is related to feature-distribution metrics such as FID~\cite{heusel2017gans} and MDFS~\cite{ni2024opinion}, and to Gaussian anomaly-detection methods such as PaDiM~\cite{defard2021padim}. Unlike these methods, SCODA does not use feature statistics as a generic image-quality prior, set-level comparison, or local anomaly detector. Instead, it builds a scene-conditioned reference model and casts quality as the deviation of a single rendered view from that scene's high-quality distribution, reframing QA as reference-aware, per-scene distributional consistency. 

Although effective, the fidelity signal does not explicitly capture the perceptual impact of artifacts. We therefore add a weakly-supervised distortion-aware realism cue and combine it with fidelity through bounded unsupervised corrections. Figure~\ref{fig:teaser} shows examples that existing QA metrics can contradict human rankings under varying reference availability, while SCODA remains perceptually consistent. Our main contributions are:
\begin{enumerate}

   \item We reformulate radiance field QA from explicit reference image comparison to scene-conditioned distributional consistency, enabling reference-aware QA of novel views without requiring a matched reference view.

    \item A lightweight weakly-supervised patch discriminator, trained to be distortion-aware, provides a complementary perceptual realism signal for localized artifacts that are not fully captured by scene-level manifold fidelity.

    \item We introduce a simple, yet effective, unsupervised fusion strategy that combines semantic fidelity with calibrated realism scores.

    \item The proposed method is not trained to regress on human score, yet demonstrates strong cross-dataset generalization across 3DGS and NeRF QA benchmark datasets.

\end{enumerate}

\section{Related Work}

\paragraph{Quality Assessment for 3D Reconstruction.}~Visual QA for 3D reconstructed scenes has gained increasing attention alongside the rapid development of 3D representations. Most existing work has focused on constructing subjectively-annotated datasets and benchmarking protocols to understand how rendering and compression choices affect perceived quality. Explicit-Nerf-QA~\cite{xing2024explicit} targets NeRF compression, classifies NeRF-induced artifacts into 9 types, and reports that FR metrics can reach correlations of around 0.85 with human mean opinion scores (MOS), whereas NR metrics perform substantially worse. Onuoha \textit{et al.}~\cite{onuoha2025benchmark} provide an annotated dataset of 16 scenes rendered using multiple NeRF models. Their analysis reports inconsistent behavior among FR metrics, indicating that conventional QA measures are not reliable for evaluating neural view synthesis.

The limitations of existing metrics become especially pronounced for 3DGS. Artifacts of GS often stem from both the rasterization pipeline and the compression of Gaussian parameters, resulting in view-dependent degradations that image-based metrics fail to capture adequately. Recent datasets have therefore been designed specifically for GS. Martin \textit{et al.}~\cite{martin2025gs} perform a subjective study across various GS methods and camera paths, while 3DGS-VBench~\cite{xing20253dgs} supports large-scale evaluation of GS-rendered videos.~These works highlight that, since GS models are rendered along trajectories, FR QA becomes impractical due to limited available references.~More recently, 3DGS-IEval-15K~\cite{xing20253dgs2} was proposed as a large-scale dataset for single-view QA of compressed 3DGS, comprising 15,200 images rendered from 10 real-world scenes using multiple compression methods. The dataset explicitly selects challenging viewpoints and reports comprehensive metric performance, highlighting the need for objective QA methods that can handle arbitrary poses, view-dependent artifacts, and diverse rendering or compression conditions.

Subjective tests provide ground-truth human judgments, but collecting such data is costly and time-consuming, motivating objective metrics that predict human opinions.

\begin{figure*}[h]
\centering

\includegraphics[width=0.85\linewidth,height=6.5cm]{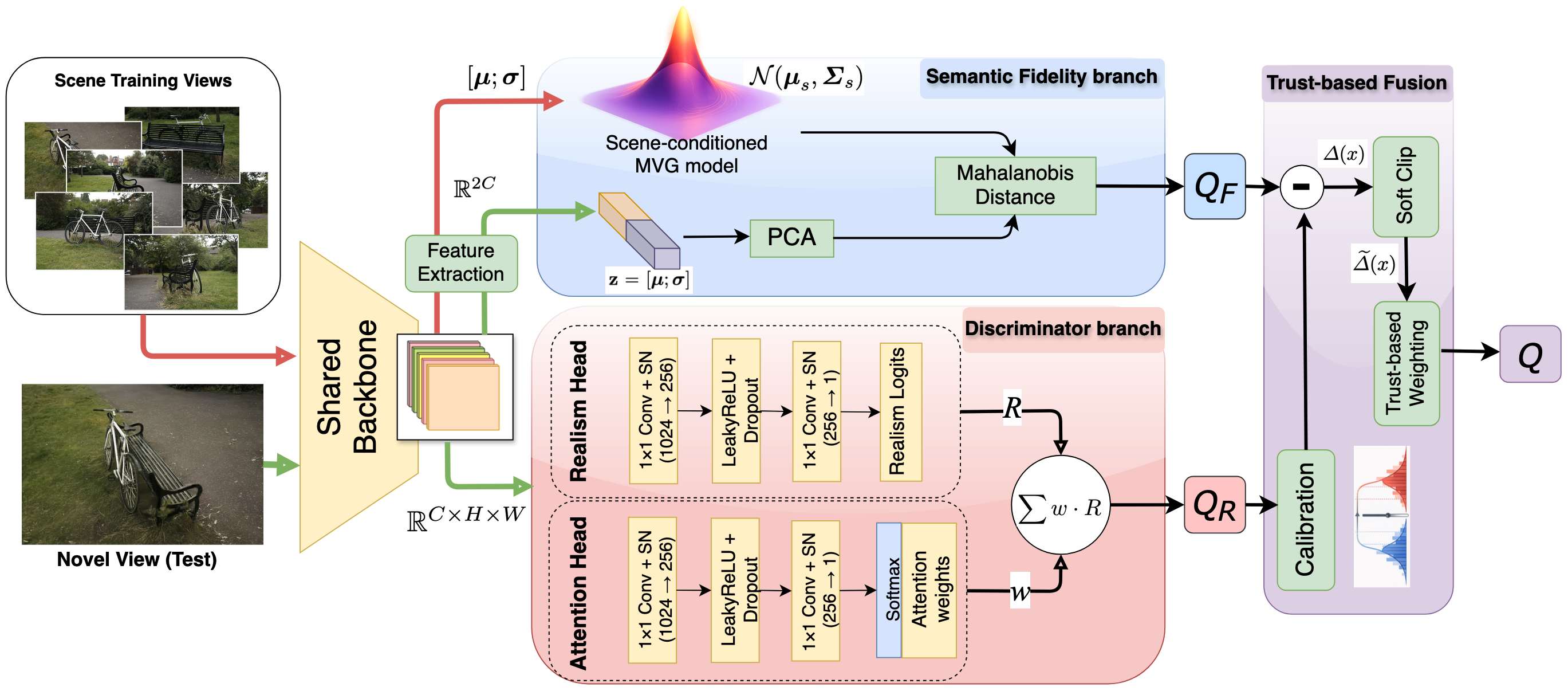}
\caption{Overview of the method. High-quality scene observations define a reference scene manifold for semantic fidelity estimation, while a distortion-aware discriminator provides a complementary realism signal. The two cues are fused to predict perceptual quality.}
\label{fig:overview}
\end{figure*}


\paragraph{Objective QA.} Objective QA for 3D scene renderings from NeRF and 3DGS have commonly relied on FR, NR, and more recently non-aligned or CR metrics.~FR metrics such as PSNR, SSIM~\cite{ssim}, LPIPS~\cite{lpips}, and DISTS~\cite{dists} are widely used when paired references are available.~However, this assumption often fails in novel-view synthesis, where renderings may come from arbitrary camera poses and exact ground-truth references may be unavailable. Recent benchmarks also show that many FR metrics misalign with human judgments on NeRF and 3DGS renderings~\cite{onuoha2025benchmark,martin2025gs,xing20253dgs2}.

NR methods predict quality directly from the rendered image. Opinion-aware NR approaches map features to human scores~\cite{chen2024topiq}, but often struggle to generalize to unseen scenes or artifacts. Conversely, opinion-unaware NR approaches leverage generic statistical regularities of natural images without using human ratings (e.g., NIQE~\cite{mittal2012making, zhang2015feature}). However, these general natural-scene priors fail to capture the complex, view-dependent distortions characteristic of modern neural scene representations. Recent NR models have been tailored specifically to neural view synthesis~\cite{qu2024nerf, qu2025nvs}, modeling spatial quality alongside angular consistency via self-supervised learning~\cite{qu2025nvs}.

Similarly, multi-view light field QA methods exploit spatial-angular representations using convolutional patches~\cite{deeblif}, handcrafted descriptors~\cite{satvblif}, attention layers~\cite{mafblif}, or frequency-domain analysis~\cite{fablfqa}. However, these methods are limited by regularly sampled view grids or narrow-baseline, and do not naturally generalize to view-dependent distortions of modern representations like 3DGS.

Recent CR methods have relaxed the strict paired-reference assumption by comparing a synthesized view with one or more views of the same scene. CrossScore~\cite{wang2024crossscore} assesses quality of a query image by comparing it with multiple unregistered scene views using a cross-attention network. PuzzleSim~\cite{hermann2025puzzle} constructs a scene-specific patch distribution from non-aligned views and uses patch-level similarity to detect local artifacts. NOVA~\cite{ghildyal_nova_2026} compares views using learned perceptual embeddings without requiring pixel-level alignment. These methods move beyond conventional FR evaluation when exact reference views are unavailable while they still rely on explicit reference views that provide sufficient overlap, nearby viewpoint coverage, or locally matchable scene content. In wide-baseline radiance-field settings, this assumption can collapse, motivating a different reference-aware formulation that does not require selecting a matched or neighboring reference image.

In summary, recent benchmarks reveal three persistent challenges: (a) \textit{Reference availability}: FR QA is often ill-defined for novel viewpoints and camera trajectories, while CR methods still require sufficient overlap or nearby views. (b) \textit{Radiance field artifacts}: local and view-dependent distortions are poorly captured by generic metrics. (c) \textit{Generalization}: existing QA metrics often degrade across scenes, datasets, and representations, especially when transferring between NeRF- and 3DGS-based renderings. NR metrics often require retraining or calibration on new datasets to remain competitive. Our proposed QA method addresses these challenges by redefining the notion of reference by building a reference scene distribution from pristine views, enabling reference-aware QA without direct image-to-image reference comparison.~Combined with a distortion-aware discriminator, the framework achieves robust performance across NeRF- and GS-based datasets.

\section{Method}

This section presents SCODA, our Scene-COnditioned and Distortion-Aware method for QA of radiance fields. SCODA combines two complementary cues including a fidelity branch measuring global consistency with the scene, and a distortion-aware realism branch. The two cues capture complementary quality information, motivating their combination in Sec.~3.3. Fig.~\ref{fig:overview} provides an overview of the method, highlighting three main components described in the following subsections. In the default setting, SCODA is weakly supervised only through the realism branch: MOS quantiles are used to form coarse clean/degraded groups for discriminator training. The fidelity branch and fusion stage are not supervised, and no component regresses MOS as a continuous target.

\subsection{Semantic Fidelity Measure}

We model perceptual image quality as the deviation of a test image from a scene-specific reference distribution in deep feature space, where degradations correspond to departures along semantically meaningful directions. Wasserstein distances have recently proven effective for measuring perceptual dissimilarity, comparing two images via the transport cost between their feature distributions \cite{qiu2024wasserstein}, and have been used as a distortion loss for perceptual optimization of neural image codecs \cite{balle2025good}. Here however, our goal is to evaluate how far a single image deviates from the distribution of high-quality observations of the same scene, framing the problem as a comparison between an observation and a population distribution.

Let $x \in \mathcal{X}$ denote an image and $\phi(\cdot)$ a pretrained deep backbone that extracts a feature representation $\mathbf{z} = \phi(x) \in \mathbb{R}^d$. We instantiate $\phi(\cdot)$ as a pretrained ResNet-50 and extract activations from the third residual block (layer3). Intermediate (layer3) features balance structural and perceptual cues, whereas earlier layers capture low-level texture and deeper layers encode semantics less sensitive to distortions (Supp. Sec. 3.2.2). Let $\mathbf{F} \in \mathbb{R}^{C \times H \times W}$ denote the resulting feature tensor. To obtain a compact descriptor, we compute the channel-wise mean and standard deviation over spatial locations $(H,W)$ and concatenate them to form $\mathbf{z} = [\boldsymbol{\mu}; \boldsymbol{\sigma}] \in \mathbb{R}^{2C}$. For dimensionality reduction and improved numerical stability, $\mathbf{z}$ can be projected onto a lower-dimensional subspace using principal component analysis (PCA) before constructing the scene-specific model.

For each scene $s$, we model the distribution of features from high-quality reference views as a compact multivariate Gaussian (MVG) in feature space (referred to as the \emph{scene manifold}) $p_s(\mathbf{z}) = \mathcal{N}(\mathbf{z} \mid \boldsymbol{\mu}_s, \boldsymbol{\Sigma}_s)$, where the parameters are estimated from $N$ pristine images
\begin{equation}
\begin{aligned}
\boldsymbol{\mu}_s
&= \frac{1}{N}\sum_{i=1}^{N}\mathbf{z}_i, \\
\boldsymbol{\Sigma}_s
&= \frac{1}{N-1}\sum_{i=1}^{N}
(\mathbf{z}_i-\boldsymbol{\mu}_s)
(\mathbf{z}_i-\boldsymbol{\mu}_s)^\top
+ \epsilon I .
\end{aligned}
\end{equation}

A small regularization $\epsilon I$ ensures numerical stability and reflects perceptual tolerance to minor feature variability. The squared 2-Wasserstein distance between two Gaussian distributions
$\mathcal{N}(\boldsymbol{\mu}_1,\boldsymbol{\Sigma}_1)$ and
$\mathcal{N}(\boldsymbol{\mu}_2,\boldsymbol{\Sigma}_2)$ is

\begin{equation}
\begin{aligned}
W_2^2
&= \|\boldsymbol{\mu}_1-\boldsymbol{\mu}_2\|_2^2 \\
&\quad + \mathrm{Tr}\!\left(
\boldsymbol{\Sigma}_1+\boldsymbol{\Sigma}_2
-2(\boldsymbol{\Sigma}_2^{1/2}
\boldsymbol{\Sigma}_1
\boldsymbol{\Sigma}_2^{1/2})^{1/2}
\right).
\end{aligned}
\end{equation}

Unlike perceptual metrics, which compare two images, we aim to measure how far a single observation deviates from a distribution of high-quality scene features. In probability theory, a single observation is represented by its empirical distribution assigning all mass to $\mathbf{z}$, i.e., the Dirac (degenerate) $\delta_{\mathbf{z}} \equiv \mathcal{N}(\mathbf{z},\mathbf{0})$. Substituting $\delta_{\mathbf{z}}$ and $\mathcal{N}(\boldsymbol{\mu}_s,\boldsymbol{\Sigma}_s)$ into (2) yields a quadratic form in $(\mathbf{z}-\boldsymbol{\mu}_s)$ up to an additive constant. Whitening the feature space using $\tilde{\mathbf{z}}=\boldsymbol{\Sigma}_s^{-1/2}(\mathbf{z}-\boldsymbol{\mu}_s)$ makes the reference distribution isotropic. Ignoring constant terms, the transport cost reduces to
\begin{equation}
    W_2^2\!\left(\delta_{z}, \mathcal{N}(\boldsymbol{\mu}_s, \boldsymbol{\Sigma}_s)\right)
    \propto
    (z - \boldsymbol{\mu}_s)^\top \boldsymbol{\Sigma}_s^{-1} (z - \boldsymbol{\mu}_s),
\end{equation}
which is precisely the Mahalanobis distance used in our metric. Consequently, the proposed semantic fidelity score $(Q_{F} = -W_2^2(.))$ is derived from the squared 2-Wasserstein formulation between a test image and the learned distribution of high-quality images for that scene. Low-quality images incur higher transport cost and therefore yield lower semantic fidelity scores. The score measures a covariance-normalized deviation from the high-quality scene distribution: directions with high natural variation across reference views are down-weighted, while deviations along stable scene-specific directions are penalized more strongly.

\subsection{Distortion-Aware Discriminator}

The MVG model measures semantic fidelity, i.e., the distance of a sample to the learned scene manifold. However, proximity to the manifold does not explicitly capture the perceptual impact of artifacts and therefore lacks distortion awareness. To better capture artifact-induced degradations, we introduce a distortion-aware patch discriminator.~Unlike the MVG model, which is obtained solely from pristine references, this model is trained to distinguish perceptually clean from degraded images using coarse pseudo-labels. The discriminator therefore learns distortion-sensitive representations complementary to semantic fidelity.

\textbf{Pseudo-labels.}~Given rendered images of a scene at different quality levels, we construct two classes of high- and low-quality images by assigning pseudo-labels using quantile-based thresholds. In our default setting, the two sets are selected using subjective mean opinion scores (MOS), which directly identify perceptually clean and degraded samples. Importantly, MOS is used only to form coarse high- and low-quality groups and the discriminator is not trained to regress, rank, or reproduce the MOS values. After thresholding, the continuous subjective scores are discarded and only binary clean/degraded labels are retained. Thus, MOS acts only as a regime selector, while the discriminator learns a patch-level realism criterion from visual evidence. We also report a fully unsupervised variant in the experiments, where coarse clean/degraded groups are selected using objective scores instead of MOS.

Let $p_{\alpha}$ denote the $\alpha$-quantile of MOS over a training set, we assign pseudo-labels as follows: samples with $MOS(x) \geq p_{0.75}$ are labeled as perceptually clean ($y=+1$), samples with $MOS(x) \leq p_{0.25}$ as degraded ($y=-1$). Samples in the intermediate range are discarded.

\textbf{Patch Discriminator.}~The discriminator follows a PatchGAN-style \cite{isola2017image} architecture built upon a pretrained ResNet-50 backbone. We extract intermediate features from the 3rd residual block (layer3), yielding a feature map $F \in \mathbb{R}^{C \times H \times W}$ with $C=1024$ channels. The backbone remains fixed during training. A lightweight convolutional head $\psi_d$ produces spatial realism logits: $R = \psi_d(F) \in \mathbb{R}^{H \times W}$. The discriminator is trained using the hinge loss
\begin{equation}
\mathcal{L}_{\text{hinge}} =
\mathbb{E}\big[\max(0, 1 - y \cdot q_R)\big],
\end{equation}
where $q_R$ is the aggregated realism score and $y \in \{-1, +1\}$ is the pseudo-label. Spectral normalization (SN) is applied to all convolutional layers in the discriminator head to stabilize optimization and control the Lipschitz constant.

\textbf{Attention-Weighted Aggregation.} Uniform averaging of patch logits assumes equal perceptual importance across spatial regions. However, distortions are often spatially localized and perceptual sensitivity varies across content. We therefore introduce attention-weighted aggregation where a second convolutional head, similar to the discriminator (with spectrally normalized $1\times1$ convolutions), produces attention logits in parallel $
A = \psi_w(F) \in \mathbb{R}^{H \times W}$. Next, spatial weights are obtained via softmax normalization:
\begin{equation}
w_{ij} =
\frac{\exp(A_{ij} / T)}
{\sum_{k,l} \exp(A_{kl} / T)},
\end{equation}
where $T$ is a temperature parameter. The realism score is computed as:
\begin{equation}
q_R = \sum_{i,j} w_{ij} R_{ij}.
\end{equation}

This formulation enables spatially adaptive perceptual pooling in which regions indicative of distortions receive higher weights, while perceptually irrelevant areas are suppressed. Both realism logits and attention weights are learned jointly end-to-end. Finally, to ensure comparability across scenes, discriminator outputs are normalized using scene-specific statistics computed from reference views:
\begin{equation}
Q_R = \frac{q_R - \mu_R^{(s)}}{\sigma_R^{(s)}},
\end{equation}
where $\mu_R^{(s)}$ and $\sigma_R^{(s)}$ denote the mean and standard deviation of discriminator logits for pristine images of scene $s$.
\subsection{Score Fusion}
\label{sec:unsup_gated_fusion}

The semantic score $Q_F$ derived from the scene-conditioned model provides a strong fidelity signal, while the realism $Q_R$ captures distortion-sensitive perceptual validity.~Since $Q_F$ reflects global distributional consistency, it can miss localized artifacts that $Q_R$ detects, motivating their fusion.

We found that a fixed linear combination of the two scores tends to blur their complementary behavior, as $Q_F$ is already a very strong signal with high accuracy. Therefore, in our unsupervised (MOS-free) fusion, $Q_F$ remains the primary estimate due to its reference-aware modeling and strong generalizability. Rather than linearly reweighting the two signals, we introduce the realism score as a bounded residual correction to $Q_F$. The correction depends on the disagreement between the two scores and is softly constrained to prevent either component from dominating.

\textbf{Calibration.} 
The two scores, $Q_F$ and $Q_R$, lie on different scales with different marginal distributions. Objective QA frameworks often align predicted scores via subjective ratings, for example using monotonic logistic regression fitted to MOS~\cite{chen2025toward}, which introduces dependence on human annotations on a specific dataset. To avoid dataset-specific score fitting for better cross-dataset generalization, we perform non-parametric quantile alignment on a small set of calibration scenes, while evaluation is conducted on held-out scenes. This alignment makes the two signals comparable before fusion, without relying on MOS supervision.

Let $\mathcal{M}_{Q_R}$ and $\mathcal{M}_{Q_F}$ denote the empirical cumulative distribution functions of ${Q_R}$ and ${Q_F}$, respectively. The aligned realism score is defined as:

\begin{equation}
\widetilde{Q}_R(x)
=
\mathcal{M}_{Q_F}^{-1}\!\big(
\mathcal{M}_{Q_R}(Q_R(x))
\big).
\end{equation}

This transformation preserves rank information while matching the marginal distribution of $\widetilde{Q}_R$ to that of $Q_F$.

\textbf{Bounded Residual Correction.} The realism cue should refine rather than override the reference-aware $Q_F$; we therefore apply a saturating correction that cannot dominate under large disagreement. The disagreement between aligned realism and fidelity scores is defined as $\Delta(x) = \widetilde{Q}_R(x) - Q_F(x)$. Rather than directly adding $\Delta(x)$, we use a bounded correction:
\begin{equation}
\widetilde{\Delta}(x)
=
S \tanh\!\left(
\frac{\Delta(x)}{S}
\right),
\end{equation}
where $S$ is defined as the $p$-th percentile (e.g., $p=75$) of $|\Delta|$ computed on calibration data. The soft clipping keeps small corrections nearly linear, $\widetilde{\Delta}(x) \approx \Delta(x)$ when $|\Delta| \ll S$, while saturating large disagreements at $\pm S$ to improve fusion stability.

\textbf{Adaptive Weighting.} We modulate the residual correction based on the quality regime indicated by the fidelity score. Extremely high or low fidelity scores often correspond to regimes where quality measurements become noisy. Moreover, the discriminator is trained using quantile-based pseudo-labels (e.g., $p_{25}$ and $p_{75}$), which encourages confident real/fake decisions in extreme regimes, therefore, realism scores in these regions may saturate. To reduce over-correction in such cases, we derive a symmetric trust weight from normalized fidelity scores:
\begin{equation}
Q_F^{\mathrm{z}}(x)
=
\frac{Q_F(x) - \mu_F}{\sigma_F},
\end{equation}
where $\mu_F$ and $\sigma_F$ are calibration statistics. We define a two-tailed trust function
\begin{equation}
w(x)
=
\frac{1}{1 + \left| Q_F^{\mathrm{z}}(x) \right|^2},
\end{equation} which reduces correction strength in extreme semantic regimes and allows maximal flexibility near the center of the fidelity distribution. Then, the final score for image $x$ is
\begin{equation}
Q(x)
=
Q_F(x)
+
w(x)\,
\widetilde{\Delta}(x).
\end{equation}

When fidelity and realism agree ($\Delta \approx 0$), the correction vanishes and the score reduces to $Q_F$. When disagreement occurs, realism information is injected in a bounded and confidence-aware manner. 

\begin{table*}[h]
\scriptsize
\centering
\caption{Quantitative comparison across 3DGS-IEval-15K, GS-QA, and NeRF QA datasets.
Rows report SRCC/PLCC/KRCC for each dataset. Best results are in \textbf{bold}, second best are \underline{underlined}. SCODA-F and SCODA-R denote the fidelity and realism branches of SCODA.}
\label{tab:iqa_results}
\setlength{\tabcolsep}{1.7pt}
\renewcommand{\arraystretch}{1.12}

\resizebox{\textwidth}{!}{
\begin{tabular}{ll|cccccccc|cccc|ccc|ccc}
\toprule
 & 
& \multicolumn{8}{c|}{\textbf{Full-reference}}
& \multicolumn{4}{c|}{\textbf{No-reference}}
& \multicolumn{3}{c|}{\textbf{Cross-reference}}
& \multicolumn{3}{c}{\textbf{Scene-conditioned}} \\
\cmidrule(lr){3-10} \cmidrule(lr){11-14} \cmidrule(lr){15-17} \cmidrule(lr){18-20}
\textbf{Dataset} & \textbf{Criteria}
& \rot{PSNR} 
& \rot{SSIM}
& \rot{VIF} 
& \rot{FSIM}
& \rot{IW-SSIM} 
& \rot{DISTS} 
& \rot{LPIPS} 
& \rot{DeepDC}
& \rot{TOPIQ} 
& \rot{MUSIQ} 
& \rot{ARNIQA} 
& \rot{LIQE}
& \rot{CrossScore} 
& \rot{PuzzleSim}
& \rot{NOVA} 
& \rot{SCODA-F} 
& \rot{SCODA-R} 
& \rot{\textbf{SCODA}} \\
\midrule

\multirow{3}{*}{\shortstack{3DGS-\\IEval-15K}}
& SRCC
& 0.645 & 0.679 & 0.559 & 0.733 & 0.694 & 0.820 & 0.731 & 0.855
& 0.495 & 0.497 & 0.371 & 0.546
& 0.490 & 0.392 & 0.795
& \underline{0.860} & 0.849 & \textbf{0.898} \\

& PLCC
& 0.639 & 0.665 & 0.569 & 0.712 & 0.692 & 0.813 & 0.732 & 0.853
& 0.508 & 0.507 & 0.377 & 0.555
& 0.492 & 0.417 & 0.795
& \underline{0.856} & 0.846 & \textbf{0.902} \\

& KRCC
& 0.457 & 0.489 & 0.393 & 0.536 & 0.506 & 0.621 & 0.537 & 0.664
& 0.342 & 0.343 & 0.250 & 0.377
& 0.337 & 0.271 & 0.596
& \underline{0.666} & 0.640 & \textbf{0.705} \\

\midrule

\multirow{3}{*}{GS-QA}
& SRCC
& 0.678 & 0.738 & 0.653 & 0.808 & 0.810 & 0.784 & 0.722 & \textbf{0.860}
& 0.640 & 0.512 & 0.762 & 0.609
& 0.700 & {0.674} & 0.805
& 0.845 & 0.733 & \underline{0.858} \\

& PLCC
& 0.717 & 0.780 & 0.693 & 0.808 & 0.830 & 0.804 & 0.731 & \textbf{0.856}
& 0.630 & 0.614 & 0.754 & 0.631
& 0.748 & {0.677} & 0.778
& 0.844 & 0.740 & \underline{0.855} \\

& KRCC
& 0.550 & 0.620 & 0.499 & \underline{0.681} & 0.679 & 0.596 & 0.523 & \textbf{0.707}
& 0.480 & 0.388 & 0.567 & 0.458
& 0.530 & {0.528} & 0.652
& 0.665 & 0.594 & 0.679 \\

\midrule

\multirow{3}{*}{NeRF QA}
& SRCC
& 0.727 & 0.640 & 0.828 & 0.712 & 0.742 & \textbf{0.909} & 0.660 & 0.871
& 0.539 & 0.515 & 0.247 & 0.633
& 0.512 & 0.612 & 0.815
& 0.866 & 0.774 & \underline{0.880} \\

& PLCC
& 0.738 & 0.636 & 0.843 & 0.713 & 0.765 & \textbf{0.911} & 0.650 & 0.873
& 0.673 & 0.620 & 0.455 & 0.708
& 0.520 & 0.630 & 0.800
& 0.870 & 0.783 & \underline{0.882} \\

& KRCC
& 0.541 & 0.484 & 0.631 & 0.521 & 0.555 & \textbf{0.732} & 0.491 & {0.690}
& 0.378 & 0.362 & 0.160 & 0.454
& 0.358 & 0.412 & 0.631
& 0.677 & 0.590 & \underline{0.693} \\

\bottomrule
\end{tabular}
}
\end{table*}

\section{Evaluation}

\textbf{Benchmark Datasets.} We evaluate the proposed model primarily on 3DGS-IEval-15K \cite{xing20253dgs2}, the largest subjectively-annotated dataset for compressed GS models.~It comprises 15,200 rendered images from 10 scenes, generated by 6 GS compression methods across 20 viewpoints per scene.~To further assess robustness, we additionally evaluate the model on two complementary datasets;~GS-QA \cite{martin2025gs} contains 64 compressed GS video sequences rendered along camera trajectories, reflecting continuous viewpoint exploration and providing insights into performance in dynamic scenarios.~Finally, the NeRF Quality Dataset \cite{onuoha2025benchmark} is used to examine cross-representation generalization.~It includes 102 stimuli from 16 synthetic and real-world scenes using four NeRF models. Instead of explicit compression modules, quality levels are controlled by varying key model and rendering parameters. For cross-dataset evaluation, the discriminator and fusion training/calibration are done on 3DGS-IEval and transferred to GS-QA and NeRF QA.

\begin{table}[t]
\centering
\scriptsize
\caption{Effect of attention and training set size on SRCC (3DGS-IEval-15K). The discriminator is trained using 3–5 scenes.}
\label{tab:attention}
\setlength{\tabcolsep}{3.pt}
\begin{tabular}{l|cc|cc|cc}

\toprule
\multirow{2}{*}{\textbf{Method}} 
& \multicolumn{2}{c|}{\textbf{3 scenes}}
& \multicolumn{2}{c|}{\textbf{4 scenes}}
& \multicolumn{2}{c}{\textbf{5 scenes}} \\
\cmidrule(lr){2-3}\cmidrule(lr){4-5}\cmidrule(lr){6-7}
& w/o Attn & w/ Attn & w/o Attn & w/ Attn & w/o Attn & w/ Attn \\
\midrule
\textbf{SCODA-R}   & 0.805 & 0.853 & 0.811 & 0.862 & 0.812 & 0.869 \\
\textbf{SCODA-F}         & 0.862 & 0.862    & 0.879 &  0.879   & 0.880 &  0.880    \\
\textbf{SCODA}  & 0.875 & 0.894 & 0.885 & \textbf{0.902} & 0.887 & \textbf{0.902} \\
\bottomrule
\end{tabular}
\end{table}

\textbf{Evaluation Criteria.} We report SRCC, PLCC, and KRCC between predicted scores and MOS. SRCC/KRCC measure ranking consistency, while PLCC measures prediction accuracy after logistic alignment~\cite{mahmoudpour2026benchmarking}. Higher values indicate better agreement with human judgments.

\textbf{Quantitative Results.}~We compare SCODA against representative FR metrics, including PSNR, SSIM~\cite{ssim}, VIF~\cite{vif2006}, FSIM~\cite{fsim}, IW-SSIM~\cite{iwssim}, DISTS~\cite{dists}, LPIPS~\cite{lpips}, and DeepDC~\cite{chen2025toward};~NR metrics, including TOPIQ~\cite{chen2024topiq}, MUSIQ~\cite{ke2021musiq}, ARNIQA~\cite{Arniqa_agnolucci2024arniqa}, and LIQE~\cite{liqe_zhang2023liqe}; and CR metrics CrossScore~\cite{wang2024crossscore}, PuzzleSim~\cite{hermann2025puzzle}, and NOVA~\cite{ghildyal_nova_2026}. NR metrics are used with their pretrained models without retraining.~Table~\ref{tab:iqa_results} compares SCODA with representative FR, NR, and CR QA methods across three benchmark datasets.~On the 3DGS-IEval-15K dataset, the fidelity branch already achieves strong correlation with human judgments, and the full model obtains the best overall correlation. This confirms that modeling high-quality observations as a reference scene distribution provides an effective scene-conditioned assessment signal.

On GS-QA, SCODA achieves results comparable to the top-performing metric DeepDC under continuous viewpoint variations and temporally coherent distortions. For frame-level aggregation, we use Minkowski temporal pooling with sensitivity parameter $p=10$~\cite{batsi2020improved}, which emphasizes low-quality views rather than uniformly averaging (Supp. Sec. 3.4). On the NeRF QA dataset, SCODA also generalizes across representation paradigms and distortion types. Although DISTS achieves the highest correlation on this dataset, SCODA remains among the strongest methods while avoiding direct comparison with an aligned explicit reference view. Note that FR metrics are evaluated under the condition that an aligned reference view is available.

\textbf{Effect of Attention and Training Size.} Out of the 10 available scenes in the 3DGS-IEval dataset, a subset is used to train the discriminator, and the same subset serves as calibration data for normalization and score alignment in the fusion stage. Evaluation is done on the remaining held-out scenes. Table \ref{tab:attention} reports SRCC results for different numbers of calibration/training scenes, and compares discriminator performance with and without attention.~As shown, introducing attention consistently improves discriminator performance. This confirms that attention-weighted aggregation directs the discriminator toward distortion-sensitive regions, yielding a more reliable assessment than uniform averaging. The results also suggest that effective learning can be achieved using only a small subset of scenes. Given the limited available training data, adopting a simple discriminator is beneficial, as it maintains stability and avoids overfitting while still providing reliable distortion detection. 
\begin{figure*}[h]
\centering

\includegraphics[width=0.88\linewidth]{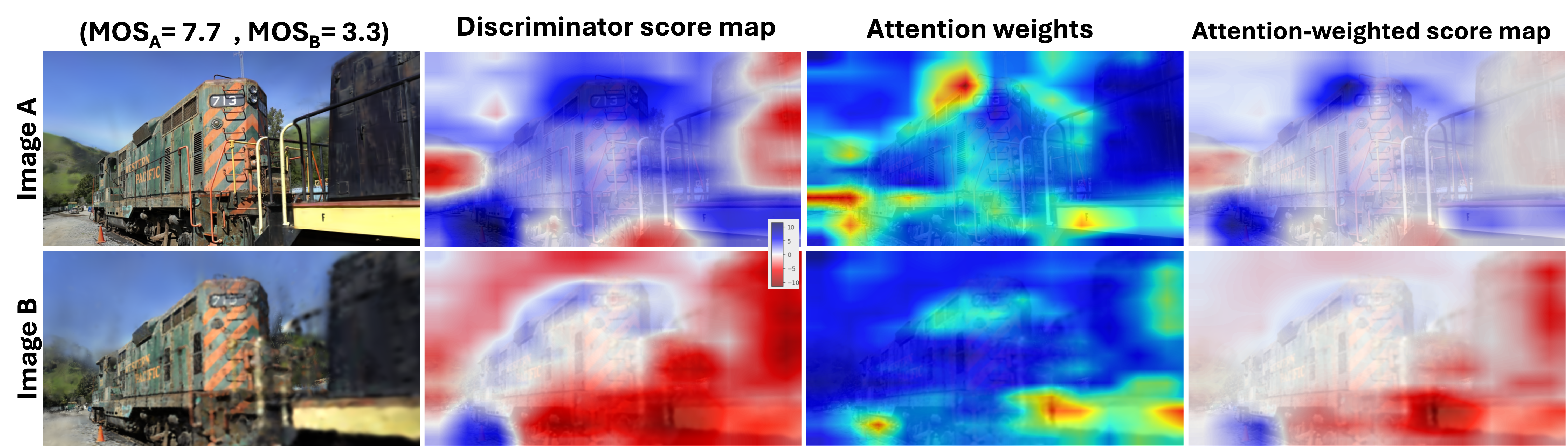}
\caption{Visualization of attention-weighted realism estimation. For two renderings at different subjective quality, we show raw patch-level realism scores, learned attention weights, and the attention-weighted response. The attention mechanism suppresses less informative regions and emphasizes salient and distortion-sensitive structures, yielding a more localized and perceptually aligned realism signal.}
\label{fig:heatmap}
\end{figure*}

\begin{figure}[h]
\centering
\includegraphics[width=0.80\linewidth]{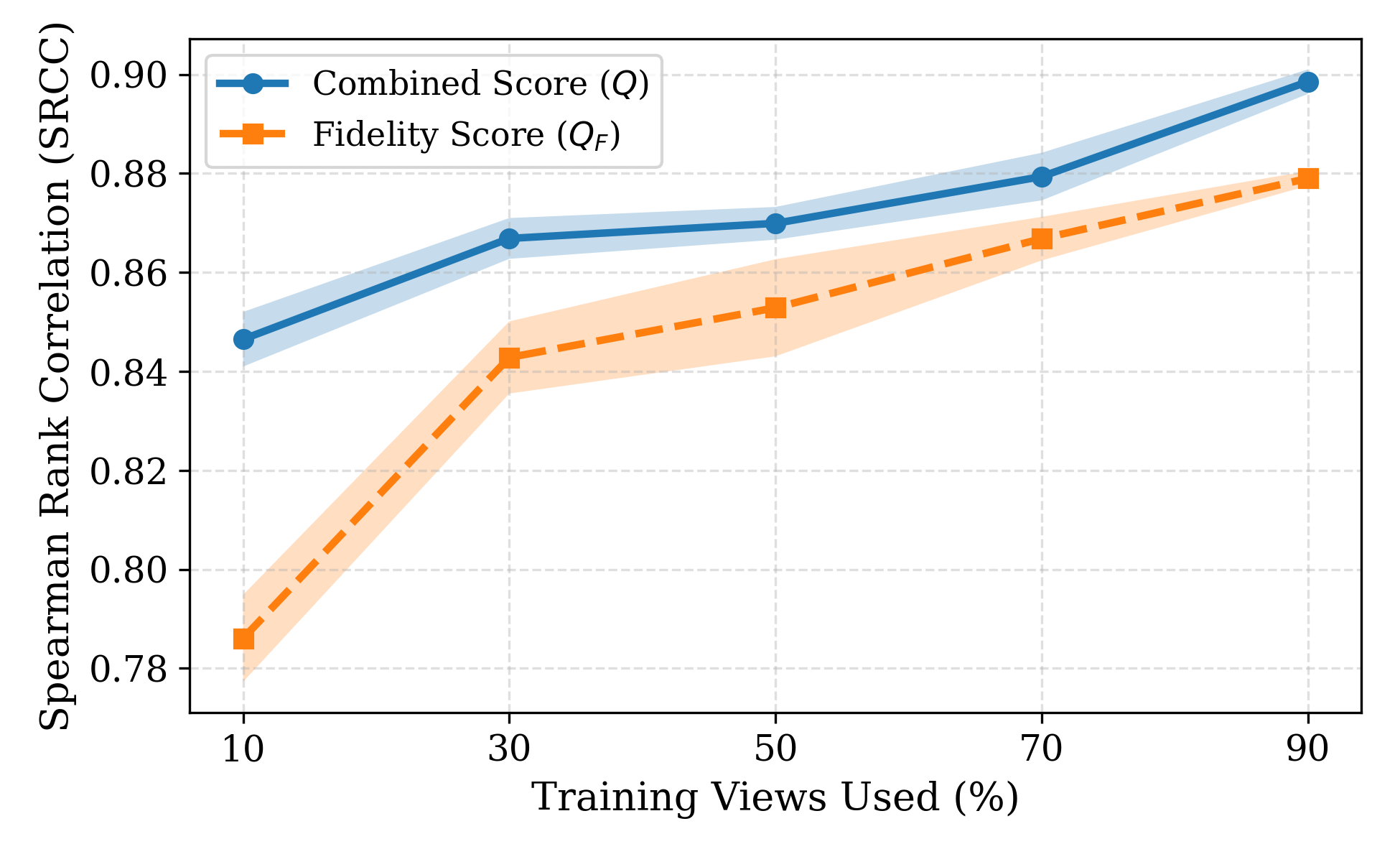}

\caption{Effect of reference-view count on metric performance.}
\label{fig:fusion}
\end{figure}

\begin{table}[t]
\centering
\footnotesize
\caption{Effect of pseudo-label source for the realism branch on the 3DGS-IEval held-out split, reported using SRCC.}
\label{tab:pseudolabel_main}
\setlength{\tabcolsep}{4pt}
\begin{tabular}{l|ccc}
\toprule
{Pseudo-label source} & {SCODA-R} &{SCODA-F} & {SCODA} \\
\midrule
{MOS quantiles (weak sup.)} & {0.869} & {0.880}& {\textbf{0.902}} \\
{$Q_F$ quantiles (MOS-free)} & {0.850} & {0.880}& {0.897}\\
{DISTS quantiles (MOS-free)} & {0.822} & {0.880}& {0.888} \\
\bottomrule
\end{tabular}
\end{table}

Fig.~\ref{fig:heatmap} visualizes the attention-based discriminator. Learned attention emphasizes distortion-sensitive regions and suppresses less informative areas, preventing localized artifacts from being diluted by uniform aggregation. The attention weights are not directly supervised, but emerge from the discriminator objective.

\vspace{-5mm}

\paragraph{Supervision Analysis for the Realism Branch.} To clarify the supervision used by the realism branch, Table~\ref{tab:pseudolabel_main} compares three pseudo-label sources for discriminator training. The default setting uses subjective scores to form coarse binary clean/degraded groups. The other variants are fully MOS-free: high/low-quality samples are selected using (a) the fidelity score $Q_F$ and (b) DISTS as an external perceptual metric to form the pseudo-labels. All variants use the same discriminator architecture and fusion rule, examined on 5 held-out scenes (7600 images). The MOS-based setting performs best, the $Q_F$ quantiles perform closest to the weak supervision, and DISTS quantiles are weaker but still improve the fused score. MOS-free variants remain competitive, while MOS quantiles give stronger results by providing human-perception-based clean/degraded regimes rather than proxy objective regimes.

\textbf{Fusion Mechanism Analysis.} Results in Table~\ref{tab:attention} imply that the fusion strategy can robustly enhance performance even when one of the signals is weaker. This behavior is further illustrated in Fig.~\ref{fig:fusion}, which shows the performance of the MVG model under different numbers of training views and analyzes the stability of the fusion mechanism. In this experiment, the MVG model is estimated using a fraction of the available reference views. Results are reported as the median over 10 random trials at each fraction. As shown in the figure, when only 10\% of the reference views are used (approximately 17–30 views per scene), the standalone manifold model exhibits a performance drop. In contrast, the final fused model remains substantially more stable, indicating that the realism signal compensates for reduced accuracy in manifold estimation.


\textbf{Backbone / PCA Ablation.}~Table~\ref{tab:backbone_ablation_full} evaluates different pretrained backbones and PCA ratios for the fidelity branch. ResNet-50 layer3 achieves the strongest performance, suggesting that intermediate CNN features are well suited for scene-conditioned modeling.~Performance also improves from 10\% to 30\% retained PCA components and then saturates, indicating that most perceptually relevant variation is captured in a lower-dimensional subspace. We also report test-time computational cost in Supp. Sec. 2. SCODA requires approximately $0.10$s per $1237 \times 822$ image on CPU, demonstrating lower inference overhead than the competing methods.

\begin{table}[t]
\centering
\footnotesize
\caption{Backbone ablation for the semantic fidelity on 3DGS-15K. SRCC is reported for PCA ratios from 10\% to 100\% features.}
\label{tab:backbone_ablation_full}
\setlength{\tabcolsep}{4pt}
\renewcommand{\arraystretch}{1.1}
\begin{tabular}{lccccc}
\toprule
\textbf{Backbone} 
& \textbf{10\%} 
& \textbf{20\%} 
& \textbf{30\%} 
& \textbf{60\%} 
& \textbf{100\%} \\
\midrule
ResNet-50 (layer3) 
& 0.691 
& 0.838 
& 0.859 
& 0.860 
& 0.859 \\

VGG16-BN (conv4\_3) 
& 0.816 
& 0.832 
& 0.842 
& 0.847 
& 0.848 \\

CLIP ViT-B/32 (patch) 
& 0.705 
& 0.749 
& 0.807 
& 0.828 
& 0.830 \\

ViT-B/16 (patch) 
& 0.601 
& 0.609 
& 0.765 
& 0.790 
& 0.795 \\
\bottomrule
\end{tabular}
\end{table}

\section{Conclusions}
We presented SCODA, a scene-conditioned framework for perceptual quality assessment of radiance-field renderings. Instead of comparing a test view with an aligned or neighboring reference image, SCODA constructs a reference scene model from high-quality observations and evaluates each rendering by its deviation from this scene-level distribution.~To improve sensitivity to artifacts, we further introduced a distortion-aware realism signal and a fusion strategy that applies bounded corrections only when the two cues disagree. Experiments across 3DGS images, GS trajectories, and NeRF-generated views show strong agreement with human judgments and robust generalization.

%
%

{
    \small
    \bibliographystyle{ieeenat_fullname}
    \bibliography{main}
}

\end{document}